# Multi-criteria neutrosophic decision making method based on score and accuracy functions under neutrosophic environment

*Rıdvan Şahin*

Department of Mathematics, Faculty of Science, Ataturk University, Erzurum, 25240, Turkey

ABSTRACT

A neutrosophic set is a more general platform, which can be used to present uncertainty, imprecise, incomplete and inconsistent. In this paper a score function and an accuracy function for single valued neutrosophic sets is firstly proposed to make the distinction between them. Then the idea is extended to interval neutrosophic sets. A multi-criteria decision making method based on the developed score-accuracy functions is established in which criterion values for alternatives are single valued neutrosophic sets and interval neutrosophic sets. In decision making process, the neutrosophic weighted aggregation operators (arithmetic and geometric average operators) are adopted to aggregate the neutrosophic information related to each alternative. Thus, we can rank all alternatives and make the selection of the best of one(s) according to the score-accuracy functions. Finally, some illustrative examples are presented to verify the developed approach and to demonstrate its practicality and effectiveness.

## 1.Introduction

The concept of neutrosophic set developed by Smarandache ([16], [17]) is a more general platform which generalizes the concept of the classic set, fuzzy set [34], intuitionistic fuzzy set [1] and interval valued intuitionistic fuzzy sets ([2],[3]). In contrast to intuitionistic fuzzy sets and also interval valued intuitionistic fuzzy sets, indeterminacy is characterized explicitly in the neutrosophic set. A neutrosophic set has three basic components such that truth membership, indeterminacy membership and falsity membership, and they are independent. However, the neutrosophic set generalizes the above mentioned sets from philosophical point of view and its functions $T_A(x)$, $I_A(x)$ and $F_A(x)$ are real standard or nonstandard subsets of $]0^-, 1^+[$ and are defined by $T_A(x): X \to ]0^-, 1^+[$, $I_A(x): X \to ]0^-, 1^+[$ and $F_A(x): X \to ]0^-, 1^+[$. That is, its components $T(x), I(x), F(x)$ are non-standard subsets included in the unitary nonstandard interval $]0^-, 1^+[$ or standard subsets included in the unitary standard interval $[0, 1]$ as in the intuitionistic fuzzy set. Furthermore, the connectors in the intuitionistic fuzzy set are only defined by $T(x)$ and $F(x)$ (i.e. truth-membership and falsity-membership), hence the indeterminacy $I(x)$ is what is left from 1, while in the neutrosophic set, they can be defined by any of them (no restriction) [16]. For example, when we ask the opinion of an expert about certain statement, he/she may say that the possibility in which the statement is true is 0.6 and the statement is false is 0.5 and the degree in which he/she is not sure is 0.2. For neutrosophic notation, it can be expressed as $x(0.6,0.2,0.5)$. For another example, suppose there are 10 voters during a voting process. Five vote "aye", two vote "blackball" and three are undecided. For neutrosophic notation, it can be expressed as $x(0.5,0.3,0.2)$. However, these expressions are beyond the scope of the intuitionistic fuzzy set. Therefore, the notion of neutrosophic set is more general and overcomes the aforementioned issues. But, a neutrosophic set will be difficult to apply in real scientific and engineering fields. Therefore, Wang et al. ([25], [26]) proposed the concepts of interval neutrosophic set INS and single valued neutrosophic set (SVNS), which are an instance of a neutrosophic set, and provided the set-theoretic operators and various properties of INSs and SVNSs, respectively. Then, SVNSs (or INSs) present uncertainty, imprecise, inconsistent and incomplete information existing in real world. Also, it would be more suitable to handle indeterminate information and inconsistent information. Majumdar et al. [11] introduced a measure of entropy of SVNSs. Ye [32] and proposed the correlation coefficients of SVNSs and developed a decision-making method under single valued neutrosophic environment. Broumi and Smarandache [14] extended this idea in INSs. Ye [33] also introduced the concept of simplified neutrosophic sets (SNSs), and applied the sets in an MCDM method using the aggregation operators of SNSs. Peng et al. [44] showed that some operations in Ye [33] may also be unrealistic. They defined the novel operations and aggregation operators and applied them to MCDM problems. Ye [30,31] proposed the similarity measures between SVNSs and INSs based on the relationship between similarity measures and distances. Şahin and Küçük [15] proposed the concept of neutrosophic subsethood based on distance measure for SVNSs.

We usually need the decision making methods because of the complex and uncertainty under the physical nature of the problems. By the multi-criteria decision making methods, we can choose the optimal alternative from multiple alternatives according to some criteria. The proposed set theories have provided the different multi-criteria decision making methods. Some authors ([7],[8],[9],[10],[18],[19],[ 23],[27]) studied on multi-criteria fuzzy decision-making methods based on intuitionistic fuzzy sets while some authors ([5],[13],[20],[21],[22],[28],[29]) proposed the multi-criteria fuzzy decision-making methods based on interval-valued intuitionistic fuzzy environment.

Xu and Yager [23] defined some geometric aggregation operators named the intuitionistic fuzzy weighted geometric operator, the intuitionistic fuzzy ordered weighted geometric operator and the intuitionistic fuzzy hybrid weighted geometric operator, and applied the intuitionistic fuzzy hybrid weighted geometric operator to a multi-criteria decision making problem under intuitionistic fuzzy environment. Then Xu [19] proposed the arithmetic aggregation operators which are arithmetic types of above mentioned ones. Xu and Chen [20] generalized the arithmetic aggregation operators to interval valued intuitionistic fuzzy such that the interval-valued intuitionistic fuzzy weighted geometric operator, the interval-valued intuitionistic fuzzy ordered weighted geometric operator and the interval-valued intuitionistic fuzzy hybrid weighted geometric operator, and applied the aggregation operators to a multi-criteria decision making problems by using the score function and accuracy function of interval-valued intuitionistic fuzzy numbers. The geometric aggregation operators for interval valued intuitionistic fuzzy sets are also proposed in [18].

But, until now there have been no many studies on multi-criteria decision making methods based on score-accuracy functions in which criterion values for alternatives are single valued neutrosophic sets or interval neutrosophic sets. Ye [30] proposed a multi-criteria decision making method for interval neutrosophic sets by means of the similarity measure between each alternative and the ideal alternative. Also, Ye [31] presented the correlation coefficient of SVNSs and the cross-entropy measure of SVNSs and applied them to single valued neutrosophic decision-making problems. Recently, Zhang et al. [6] established two interval neutrosophic aggregation operators such as interval neutrosophic weighted arithmetic operator and interval neutrosophic weighted geometric operator and presented a method for multi-criteria decision making problems based on the aggregation operators. Therefore the main purposes of this paper were (1) to define two measurement functions such that score function and accuracy function to rank single valued neutrosophic numbers and extend the idea in interval neutrosophic numbers, (2) to establish a multi-criteria decision making method by use of the proposed functions and neutrosophic aggregation operators for neutrosophic sets, and (3) to demonstrate the application and effectiveness of the developed methods by some numerical examples.

This paper is organized as follows. The definitions of neutrosophic sets, single valued neutrosophic sets, interval neutrosophic sets and some basic operators on them as well as arithmetic and geometric aggregation operators are briefly introduced in section 2. In section 3, the score function and the accuracy function for single valued neutrosophic numbers are introduced and studied by giving illustrative properties. Also the concepts is extended to interval neutrosophic sets in section 4. This is followed by applications of the proposed this functions to multi-criteria decision making problems in Section 5. The section 6 includes a comparison analyze. This paper is concluded in Section 7.

## 2. Preliminaries

In the following we give a brief review of some preliminaries.

*2.1 Neutrosophic set*

**Definition 2.1** [16] Let $X$ be a space of points (objects) and $x \in X$. A neutrosophic set $A$ in $X$ is defined by a truth-membership function $T_A(x)$, an indeterminacy-membership function $I_A(x)$ and a falsity-membership function $F_A(x)$. $T_A(x)$, $I_A(x)$ and $F_A(x)$ are real standard or real nonstandard subsets of $]0^-, 1^+[$. That is $T_A(x): X \to ]0^-, 1^+[$, $T_A(x): X \to ]0^-, 1^+[$ and $T_A(x): X \to ]0^-, 1^+[$. There is not restriction on the sum of $T_A(x)$, $I_A(x)$ and $F_A(x)$, so $0^- \leq \sup T_A(x) \leq \sup I_A(x) \leq \sup F_A(x) \leq 3^+$.

**Definition 2.2** [17] The complement of a neutrosophic set $A$ is denoted by $A^c$ and is defined as $T_A^c(x) = \{1^+\} \ominus T_A(x)$, $I_A^c(x) = \{1^+\} \ominus I_A(x)$ and $F_A^c(x) = \{1^+\} \ominus F_A(x)$ for all $x \in X$.

**Definition 2.3** [17] A neutrosophic set $A$ is contained in the other neutrosophic set $B$, $A \subseteq B$ iff $\inf T_A(x) \leq \inf T_B(x)$, $\sup T_A(x) \leq \sup T_B(x)$, $\inf I_A(x) \geq \inf I_B(x)$, $\sup I_A(x) \geq \sup I_B(x)$ and $\inf F_A(x) \geq \inf F_B(x)$, $\sup F_A(x) \geq \sup F_B(x)$ for all $x \in X$.

In the following, we adopt the representations $u_A(x)$, $w_A(x)$ and $v_A(x)$ instead of $T_A(x)$, $I_A(x)$ and $F_A(x)$, respectively.

*2.2 Single valued neutrosophic sets*

A single valued neutrosophic set has been defined in [25] as follows:

**Definition 2.4** [25] Let $X$ be a universe of discourse. A single valued neutrosophic set $A$ over $X$ is an object having the form

$$A = \{\langle x, u_A(x), w_A(x), v_A(x)\rangle : x \in X\}$$

where $u_A(x): X \to [0,1]$, $w_A(x): X \to [0,1]$ and $v_A(x): X \to [0,1]$ with $0 \leq u_A(x) + w_A(x) + v_A(x) \leq 3$ for all $x \in X$. The intervals $u_A(x), w_A(x)$ and $v_A(x)$ denote the truth- membership degree, the indeterminacy-membership degree and the falsity membership degree of $x$ to $A$, respectively.

**Definition 2.5** [25] The complement of an SVNS $A$ is denoted by $A^c$ and is defined as $u_A^c(x) = v(x)$, $w_A^c(x) = 1 - w_A(x)$, and $v_A^c(x) = u(x)$ for all $x \in X$. That is,
$$A^c = \{\langle x, v_A(x), 1 - w_A(x), u_A(x)\rangle : x \in X\}.$$

**Definition 2.6** [25] A single valued neutrosophic set $A$ is contained in the other SVNS $B$, $A \subseteq B$, iff $u_A(x) \leq u_B(x)$, $w_A(x) \geq w_B(x)$ and $v_A(x) \geq v_B(x)$ for all $x \in X$.

**Definition 2.7** [25] Two SVNSs $A$ and $B$ are equal, written as $A = B$, iff $A \subseteq B$ and $B \subseteq A$.

We will denote the set of all the SVNSs in $X$ by SVNS($X$). A SVNS value is denoted by $A = (a, b, c)$ for convenience.

Based on the study given in [6], we define two weighted aggregation operators related to SVNSs as follows:

**Definition 2.8** Let $A_k$ ($k = 1,2,...,n$) $\in$ SVNS($X$). The single valued neutrosophic weighted average operator is defined by
$$F_\omega = (A_1, A_2, ..., A_n) = \sum_{k=1}^{n} \omega_k A_k$$
$$= \left(1 - \prod_{k=1}^{n}\left(1 - u_{A_k}(x)\right)^{\omega_k},\right.$$
$$\left.\prod_{k=1}^{n}\left(w_{A_k}(x)\right)^{\omega_k}, \prod_{k=1}^{n}\left(v_{A_k}(x)\right)^{\omega_k}\right) \quad (1)$$

where $\omega_k$ is the weight of $A_k$ ($k = 1,2,...,n$), $\omega_k \in [0,1]$ and $\sum_{k=1}^{n} \omega_k = 1$. Especially, assume $\omega_k = 1/n$ ($k = 1,2,...,n$), then $F_\omega$ is called an arithmetic average operator for SVNSs.

Similarly, we can define the single valued neutrosophic weighted geometric average operator as follows:

**Definition 2.9** Let $A_k$ ($k = 1,2,...,n$) $\in$ SVNS($X$). The single valued neutrosophic weighted geometric average operator is defined by
$$G_\omega = (A_1, A_2, ..., A_n) = \prod_{k=1}^{n} A_k^{\omega_k}$$
$$= \left(\prod_{k=1}^{n}\left(u_{A_k}(x)\right)^{\omega_k}, 1 - \prod_{k=1}^{n}\left(1 - w_{A_k}(x)\right)^{\omega_k},\right.$$
$$\left.1 - \prod_{k=1}^{n}\left(1 - v_{A_k}(x)\right)^{\omega_k}\right) \quad (2)$$

where $\omega_k$ is the weight of $A_k$ ($k = 1,2,...,n$), $\omega_k \in [0,1]$ and $\sum_{k=1}^{n} \omega_k = 1$. Especially, assume $\omega_k = 1/n$ ($k = 1,2,...,n$), then $G_\omega$ is called a geometric average for SVNSs.

The aggregation results $F_\omega$ and $G_\omega$ are still SVNSs. Obviously, there are different emphasis points between *Definitions 2.8* and *2.9*. The weighted arithmetic average operator indicates the group's influence, so it is not very sensitive to $A_k$ ($k = 1,2,...,n$) $\in$ SVNS($X$), whereas the weighted geometric average operator indicates the individual influence, so it is more sensitive to $A_k$ ($k = 1,2,...,n$) $\in$ SVNS($X$).

**Definition 2.10** Let $A$ be a single valued neutrosophic set over $X$.

(i) A single valued neutrosophic set over $X$ is empty, denoted by $\tilde{A}$ if $u_A(x) = 1$, $w_A(x) = 0$ and $v_A(x) = 0$ for all $x \in X$.
(ii) A single valued neutrosophic set over $X$ is absolute, denoted by $\Phi$ if $u_A(x) = 0$, $w_A(x) = 1$ and $v_A(x) = 1$ for all $x \in X$.

### 2.3 Interval neutrosophic sets

An INS is an instance of a neutrosophic set, which can be used in real scientific and engineering applications. In the following, we introduce the definition of an INS.

**Definition 2.11** [26] Let $X$ be a space of points (objects) and Int[0,1] be the set of all closed subsets of [0,1]. An INS $\tilde{A}$ in $X$ is defined with the form
$$\tilde{A} = \{\langle x, u_{\tilde{A}}(x), w_{\tilde{A}}(x), v_{\tilde{A}}(x)\rangle : x \in X\}$$

where $u_{\tilde{A}}(x): X \to \text{int}[0,1]$, $w_{\tilde{A}}(x): X \to \text{int}[0,1]$ and $v_{\tilde{A}}(x): X \to \text{int}[0,1]$ with $0 \leq \sup u_{\tilde{A}}(x) + \sup w_{\tilde{A}}(x) + \sup v_{\tilde{A}}(x) \leq 3$ for all $x \in X$. The intervals $u_{\tilde{A}}(x), w_{\tilde{A}}(x)$ and $v_{\tilde{A}}(x)$ denote the truth-membership degree, the indeterminacy-membership degree and the falsity membership degree of $x$ to $\tilde{A}$, respectively.

For convenience, if let $u_{\tilde{A}}(x) = [u_{\tilde{A}}^-(x), u_{\tilde{A}}^+(x)]$, $w_{\tilde{A}}(x) = [w_{\tilde{A}}^-(x), w_{\tilde{A}}^+(x)]$ and $v(x) = [v_{\tilde{A}}^-(x), v_{\tilde{A}}^+(x)]$, then
$$\tilde{A} = \{\langle x, [u_{\tilde{A}}^-(x), u_{\tilde{A}}^+(x)], [w_{\tilde{A}}^-(x), w_{\tilde{A}}^+(x)], [v_{\tilde{A}}^-(x), v_{\tilde{A}}^+(x)]\rangle : x \in X\}$$

with the condition, $0 \leq \sup u_{\tilde{A}}^+(x) + \sup w_{\tilde{A}}^+(x) + \sup v_{\tilde{A}}^+(x) \leq 3$ for all $x \in X$. Here, we only consider the sub-unitary interval of [0,1]. Therefore, an INS is clearly neutrosophic set.

**Definition 2.12** [26] The complement of an INS $\tilde{A}$ is denoted by $\tilde{A}^c$ and is defined as $u_{\tilde{A}}^c(x) = v(x)$, $(w_{\tilde{A}}^-)^c(x) = 1 - w_{\tilde{A}}^+(x)$, $(w_{\tilde{A}}^+)^c(x) = 1 - w_{\tilde{A}}^-(x)$ and $v_{\tilde{A}}^c(x) = u(x)$ for all $x \in X$. That is,
$$\tilde{A}^c = \{\langle x, [v_{\tilde{A}}^-(x), v_{\tilde{A}}^+(x)], [1 - w_{\tilde{A}}^+(x), 1 - w_{\tilde{A}}^-(x)], [u_{\tilde{A}}^-(x), u_{\tilde{A}}^+(x)]\rangle : x \in X\}.$$

**Definition 2.13** [26] An interval neutrosophic set $\tilde{A}$ is contained in the other INS $\tilde{B}$, $\tilde{A} \subseteq \tilde{B}$, iff $u_{\tilde{A}}^-(x) \leq u_{\tilde{B}}^-(x)$, $u_{\tilde{A}}^+(x) \leq u_{\tilde{B}}^+(x)$, $w_{\tilde{A}}^-(x) \geq w_{\tilde{B}}^-(x)$, $w_{\tilde{A}}^+(x) \geq w_{\tilde{B}}^+(x)$ and $v_{\tilde{A}}^-(x) \geq v_{\tilde{B}}^-(x)$, $v_{\tilde{A}}^+(x) \geq v_{\tilde{B}}^+(x)$ for all $x \in X$.

**Definition 2.14** [26] Two INSs $\tilde{A}$ and $B$ are equal, written as $\tilde{A} = \tilde{B}$, iff $\tilde{A} \subseteq \tilde{B}$ and $\tilde{B} \subseteq \tilde{A}$.

We will denote the set of all the INSs in $X$ by $\text{INS}(X)$. An INS value is denoted by $\tilde{A} = ([a,b],[c,d],[e,f])$ for convenience.

Next, we give two weighted aggregation operators related to INSs.

**Definition 2.15** [6] Let $\tilde{A}_k$ $(k = 1,2, \ldots, n) \in \text{INS}(X)$. The interval neutrosophic weighted average operator is defined by

$$F_\omega = (\tilde{A}_1, \tilde{A}_2, \ldots, \tilde{A}_n) = \sum_{k=1}^n \omega_k \tilde{A}_k$$
$$= \left( \left[ 1 - \prod_{k=1}^n \left(1 - u_{\tilde{A}_k}^-(x)\right)^{\omega_k}, 1 - \prod_{k=1}^n \left(1 - u_{\tilde{A}_k}^+(x)\right)^{\omega_k} \right],\right.$$
$$\left[ \prod_{k=1}^n \left(w_{\tilde{A}_k}^-(x)\right)^{\omega_k}, \prod_{k=1}^n \left(w_{\tilde{A}_k}^+(x)\right)^{\omega_k} \right],$$
$$\left.\left[ \prod_{k=1}^n \left(v_{\tilde{A}_k}^-(x)\right)^{\omega_k}, \prod_{k=1}^n \left(v_{\tilde{A}_k}^+(x)\right)^{\omega_k} \right] \right) \quad (3)$$

where $\omega_k$ is the weight of $\tilde{A}_k$ $(k = 1,2, \ldots, n)$, $\omega_k \in [0,1]$ and $\sum_{k=1}^n \omega_k = 1$. Especially, assume $\omega_k = 1/n$ $(k = 1,2, \ldots, n)$, then $F_\omega$ is called an arithmetic average operator for INSs.

**Definition 2.16** [6] Let $\tilde{A}_k$ $(k = 1,2, \ldots, n) \in \text{INS}(X)$. The interval neutrosophic weighted geometric average operator is defined by

$$G_\omega = (\tilde{A}_1, \tilde{A}_2, \ldots, \tilde{A}_n) = \prod_{k=1}^n A_k^{\omega_k}$$
$$= \left( \left[ \prod_{k=1}^n \left(u_{\tilde{A}_k}^-(x)\right)^{\omega_k}, \prod_{k=1}^n \left(u_{\tilde{A}_k}^+(x)\right)^{\omega_k} \right],\right.$$
$$\left[ 1 - \prod_{k=1}^n \left(1 - w_{\tilde{A}_k}^-(x)\right)^{\omega_k}, 1 - \prod_{k=1}^n \left(1 - w_{\tilde{A}_k}^+(x)\right)^{\omega_k} \right],$$
$$\left.\left[ 1 - \prod_{k=1}^n \left(1 - v_{\tilde{A}_k}^-(x)\right)^{\omega_k}, 1 - \prod_{k=1}^n \left(1 - v_{\tilde{A}_k}^+(x)\right)^{\omega_k} \right] \right) \quad (4)$$

where $\omega_k$ is the weight of $\tilde{A}_k$ $(k = 1,2, \ldots, n)$, $\omega_k \in [0,1]$ and $\sum_{k=1}^n \omega_k = 1$. Especially, assume $\omega_k = 1/n$ $(k = 1,2, \ldots, n)$, then $G_\omega$ is called a geometric average for INSs.

The aggregation results $F_\omega$ and $G_\omega$ are still INSs. Obviously, there are different emphasis points between *Definitions 2.15* and *2.16*. The weighted arithmetic average operator indicates the group's influence, so it is not very sensitive to $\tilde{A}_k$ $(k = 1,2, \ldots, n) \in \text{INS}(X)$, whereas the weighted geometric average operator indicates the individual influence, so it is more sensitive to $\tilde{A}_k$ $(k = 1,2, \ldots, n) \in \text{INS}(X)$.

**Definition 2.17** [26] Let $A$ be an interval neutrosophic set over $X$.

(i) An interval neutrosophic set over $X$ is empty, denoted by $\tilde{A}$ if $u_{\tilde{A}}(x) = [1,1]$, $w_{\tilde{A}}(x) = [0,0]$ and $v_{\tilde{A}}(x) = [0,0]$ for all $x \in X$.
(ii) An interval neutrosophic set over $X$ is absolute, denoted by $\Phi$ if $u_{\tilde{A}}(x) = [0,0]$, $w_{\tilde{A}}(x) = [1,1]$ and $v_{\tilde{A}}(x) = [1,1]$ for all $x \in X$.

## 3. Ranking by score function

In the following, we introduce a score function for ranking SVN numbers by taking into account the truth-membership degree, indeterminacy-membership degree and falsity membership degree of SVNSs (and INSs), and discuss some basic properties.

**Definition 3.18** Let $A = (a, b, c)$ be a single valued neutrosophic number, a score function $K$ of a single valued neutrosophic value, based on the truth-membership degree, indeterminacy-membership degree and falsity membership degree is defined by

$$K(A) = \frac{1 + a - 2b - c}{2} \quad (5)$$

where $K(A) \in [-1,1]$.

The score function $K$ is reduced the score function proposed by Li ([8]) if $b = 0$ and $a + c \leq 1$.

It is clear that if truth-membership degree $a$ is bigger, and the indeterminacy-membership degree $b$ and falsity membership degree $c$ are smaller, then the score value of the SVNN $A$ is greater.

We give the following example.

**Example 3.19** Let $A_1 = (0.5, 0.2, 0.6)$ and $A_2 = (0.6, 0.4, 0.2)$ be two single valued neutrosophic values for two alternatives. Then, by applying *Definition 3.18*, we can obtain

$$K(A_1) = \frac{1 + 0.5 - 2 \times 0.2 - 0.6}{2} = 0.25$$
$$K(A_2) = \frac{1 + 0.6 - 2 \times 0.4 - 0.2}{2} = 0.3.$$

In this case, we can say that alternative $A_2$ is better than $A_1$.

**Proposition 3.20** Let $A = (a, b, c)$ be a single valued neutrosophic value. Then the score function $K$ has some properties as follows:

(i) $K(A) = 0$ if and only if $a = 2b + c - 1$.
(ii) $K(A) = 1$ if and only if $a = 2b + c + 1$.
(iii) $K(A) = -1$ if and only if $a = 2b + c - 3$.

Moreover, we have that $K(\tilde{A}) = 1$, which $\tilde{A}$ is the absolute single valued neutrosophic value, and $K(\Phi) = -1$, which $\Phi$ is the null single valued neutrosophic value.

**Theorem 3.21** Let $A_1 = (a_1, b_1, c_1)$ and $A_2 = (a_2, b_2, c_2)$ be two single valued neutrosophic sets. If $A_1 \subseteq A_2$, then $K(A_1) \leq K(A_2)$.

**Proof.** By Definition 3.18, we have that $K(A_1) = \frac{1+a_1-2b_1-c_1}{2}$ and $K(A_2) = \frac{1+a_2-2b_2-c_2}{2}$. Now, $K(A_2) - K(A_1) = ((a_2 - a_1) + 2(b_1 - b_2) + (c_1 - c_2))/2$. Since $A_1 \subseteq A_2$, $a_1 \leq a_2$, $b_1 \geq b_2$, $c_1 \geq c_2$ and hence $(a_2 - a_1) \geq 0$, $(b_1 - b_2) \geq 0$ and $(c_1 - c_2) \geq 0$. Then it follows that $K(A_2) - K(A_1) \geq 0$.

Now, we define a score function for the ranking order of the interval neutrosophic numbers (INSs).

**Definition 3.22** Let $\tilde{A} = ([a,b], [c,d], [e,f])$ be an interval neutrosophic number, a score function $L$ of an interval neutrosophic value, based on the truth-membership degree, indeterminacy-membership degree and falsity membership degree is defined by

$$L(\tilde{A}) = \frac{2 + a + b - 2c - 2d - e - f}{4} \quad (6)$$

where $L(\tilde{A}) \in [-1, 1]$.

We give the following example.

**Example 3.23** Let $\tilde{A}_1 = ([0.6, 0.4], [0.3, 0.1], [0.1, 0.3])$ and $\tilde{A}_2 = ([0.1, 0.6], [0.2, 0.3], [0.1, 0.4])$ be two interval neutrosophic values for two alternatives. Then, by applying *Definition 3.22*, we can obtain

$$L(\tilde{A}_1) = \frac{2 + 0.6 + 0.4 - 2 \times 0.3 - 2 \times 0.1 - 0.1 - 0.3}{4}$$
$$= 0.45,$$

$$L(\tilde{A}_2) = \frac{2 + 0.1 + 0.6 - 2 \times 0.2 - 2 \times 0.3 - 0.1 - 0.3}{4}$$
$$= 0.32.$$

In this case we can say that alternative $A_1$ is better than $A_2$.

**Proposition 3.24** Let $\tilde{A} = ([a,b], [c,d], [e,f])$ be an interval neutrosophic value. Then the score function $L$ has some properties as follows:

(i) $L(\tilde{A}) = 0$ if and only if $a + b = 2b + 2d + e + f - 2$.
(ii) $L(\tilde{A}) = 1$ if and only if $a + b = 2b + 2d + e + f + 2$.
(iii) $L(\tilde{A}) = -1$ if and only if $a + b = 2b + 2d + e + f - 6$.

Moreover, we have that $L(\tilde{A}) = 1$, which $\tilde{A}$ is the absolute interval neutrosophic value, and $L(\Phi) = -1$, which $\Phi$ is the null interval neutrosophic value.

**Theorem 3.25** Let $\tilde{A}_1 = ([a_1, b_1], [c_1, d_1], [e_1, f_1])$ and $\tilde{A}_2 = ([a_2, b_2], [c_2, d_2], [e_2, f_2])$ be two interval neutrosophic sets. If $\tilde{A}_1 \subseteq \tilde{A}_2$, then $L(\tilde{A}_1) \leq L(\tilde{A}_2)$.

**Proof.** By Definition 3.22, we have $L(\tilde{A}_1) = \frac{2+a_1+b_1-2c_1-2d_1-e_1-f_1}{4}$ and $L(\tilde{A}_2) = \frac{2+a_2+b_2-2c_2-2d_2-e_2-f_2}{4}$.
Now, $L(\tilde{A}_2) - L(\tilde{A}_1) = (a_2 - a_1) + (b_2 - b_1) + 2(c_1 - c_2) + 2(d_1 - d_2) + (e_1 - e_2) + (d_1 - d_2)$. Since $\tilde{A}_1 \subseteq \tilde{A}_2$, $a_1 \leq a_2$, $b_1 \leq b_2$, $c_1 \geq c_2$, $d_1 \geq d_2$ and $e_1 \geq e_2$, $f_1 \geq f_2$ and hence $(a_2 - a_1) \geq 0$, $(b_2 - b_1) \geq 0$, $(c_1 - c_2) \geq 0$, $(d_1 - d_2) \geq 0$, $(e_1 - e_2) \geq 0$ and $(f_1 - f_2) \geq 0$. Then it follows that $L(\tilde{A}_2) - L(\tilde{A}_1) \geq 0$.

## 4. Ranking by accuracy function

**Definition 4.26** Let $A = (a, b, c)$ be a single valued neutrosophic number, an accuracy function $M$ of a single valued neutrosophic value, based on the truth-membership degree, indeterminacy-membership degree and falsity membership degree is defined by

$$M(A) = a - b(1 - a) - c(1 - b) \quad (7)$$

where $M(A) \in [-1, 1]$.

**Example 4.27** Let $A_1 = (0.5, 0.2, 0.6)$ and $A_2 = (0.6, 0.4, 0.2)$ be two single valued neutrosophic values for two alternatives. Then, by applying *Definition 4.26*, we can obtain $M(A_1) = -0.08$ and $M(A_2) = 0{,}32$.

In this case, we can say that alternative $A_2$ is better than $A_1$.

Now, we extend the concept of accuracy function to interval neutrosophic numbers.

**Definition 4.28** Let $A = ([a, b], [c, d], [e, f])$ be an interval neutrosophic number. Then an accuracy function $N$ of an interval neutrosophic value, based on the truth-membership degree, indeterminacy-membership degree and falsity membership degree is defined by

$$N(A) = \frac{1}{2}(a + b - d(1 - b) - c(1 - a)$$
$$-f(1 - c) - e(1 - d)) \quad (8)$$

where $L(A) \in [-1, 1]$.

The accuracy function $N$ is reduced the accuracy function proposed by Nayagam et al. ([13]) if $c, d = 0$ and $b + f \leq 1$.

**Example 4.29** Let $\tilde{A}_1 = ([0.6, 0.4], [0.3, 0.1], [0.1, 0.3])$ and $\tilde{A}_2 = ([0.1, 0.6], [0.2, 0.3], [0.1, 0.4])$ be two interval neutrosophic values for two alternatives. Then, by applying *Definition 4.28*, we can obtain $M(A_1) = 0{,}26$ and $M(A_2) = 0{,}34$.

In this case we can say that alternative $A_2$ is better than $A_1$.

According to score and accuracy functions for SVNNs, we can obtain the following definitions.

**Definition 4.30** Suppose that $A_1 = (a_1, b_1, c_1)$ and $A_2 = (a_2, b_2, c_2)$ are two single valued neutrosophic number. Then we define the ranking method as follows:

(i) If $K(A_1) > K(A_2)$, then $A_1 > A_2$.
(ii) If $K(A_1) = K(A_2)$ and $L(A_1) > L(A_2)$, then $A_1 > A_2$.

**Definition 4.31** Suppose that $\tilde{A}_1 = ([a_1, b_1], [c_1, d_1], [e_1, f_1])$ and $\tilde{A}_2 = ([a_2, b_2], [c_2, d_2], [e_2, f_2])$ are two interval neutrosophic sets Then we define the ranking method as follows:

(i) If $K(\tilde{A}_1) > K(\tilde{A}_2)$, then $\tilde{A}_1 > \tilde{A}_2$.
(ii) If $K(\tilde{A}_1) = K(\tilde{A}_2)$ and $L(\tilde{A}_1) > L(\tilde{A}_2)$, then $\tilde{A}_1 > \tilde{A}_2$.

**Example 4.32** Let $A_1 = (0.5, 0.2, 0.6)$ and $A_2 = (0.6, 0.4, 0.2)$ be two single valued neutrosophic values for two alternatives. Then, by applying *Definition 3.18*, we can obtain $K(A_1) = K(A_2) = 0.6$ and $L(A_1) = 0.26$, $L(A_2) = -0.16$. Then it implies that $A_1 > A_2$.

From the above analysis, we develop a method based on the score function $K$ and the accuracy function $L$ for multi criteria decision making problem, which are criterion values for alternatives are the single valued neutrosophic value and the interval neutrosophic value, and define it as follows.

## 5. Multi-criteria neutrosophic decision-making method based on the score-accuracy function

Here, we propose a method for multi-criteria neutrosophic decision making problems with weights.

Suppose that $A = \{A_1, A_2, \ldots, A_m\}$ be the set of alternatives and $C = \{C_1, C_2, \ldots, C_n\}$ be a set of criteria. Suppose that the weight of the criterion $C_s$ ($s = 1,2,\ldots,n$), stated by the decision-maker, is $\omega_s$, $\omega_s \in [0,1]$ and $\sum_{s=1}^{n} \omega_s = 1$. Thus, the characteristic of the alternative $A_k$ ($k = 1,2,\ldots,m$) is introduced by the following SVNS and INS, respectively:

**Method 1**

$$A_k = \{\langle C_s, u_{A_k}(C_s), w_{A_k}(C_s), v_{A_k}(C_s)\rangle : C_s \in C\}$$

where $0 \le u_{A_k}(C_s) + w_{A_k}(C_s) + v_{A_k}(C_s) \le 3$, $u_{A_k}(C_s) \ge 0$, $w_{A_k}(C_s) \ge 0$, $v_{A_k}(C_s) \ge 0$, $s = 1,2,\ldots,n$ and $k = 1,2,\ldots,m$. The SVNS value that is the triple of values for $C_s$ is denoted by $\alpha_{ks} = (a_{ks}, b_{ks}, c_{ks})$, where $a_{ks}$ indicates the degree that the alternative $A_k$ satisfies the criterion $C_s$ and $b_{ks}$ indicates the degree that the alternative $A_k$ is indeterminacy on the criterion $C_s$, where as $c_{ks}$ indicates the degree that the alternative $A_k$ does not satisfy the criterion $C_s$ given by the decision-maker. So we can express a decision matrix $= (\alpha_{ks})_{m \times n}$. The aggregating single valued neutrosophic number $\alpha_k$ for $A_k$ ($k = 1,2,\ldots,m$) is $\alpha_k = (a_k, b_k, c_k) = F_{k\omega}(A_{k1}, A_{k2}, \ldots, A_{kn})$ or $\alpha_k = (a_k, b_k, c_k) = G_{k\omega}(A_{k1}, A_{k2}, \ldots, A)$, which is obtained by applying *Definition 2.8* or *Definition 2.9* according to each row in the decision matrix.

We can summarize the procedure of proposed method as follows:

**Step (1)** Obtain the weighted arithmetic average values by using Eq. (1) or the weighted geometric average values by Eq. (2)

**Step (2)** Obtain the score (or accuracy) $K(A_k)$ of single valued neutrosophic value $\alpha_k$ ($k = 1,2,\ldots,m$) by using Eq. (5).

**Step (3)** Rank the alternative $A_k = (k = 1,2,\ldots,m)$ and choose the best one(s) according to $(\alpha_k)$ ($k = 1,2,\ldots,m$).

**Method 2**

$$\tilde{A}_k = \{\langle C_s, [u_{\tilde{A}_k}^-(C_s), u_{\tilde{A}_k}^+(C_s)], [w_{\tilde{A}_k}^-(C_s), w_{\tilde{A}_k}^+(C_s)],$$
$$[v_{\tilde{A}_k}^-(C_s), v_{\tilde{A}_k}^+(C_s)]\rangle : C_s \in C\}$$

where $0 \le u_{\tilde{A}_k}^+(C_s) + w_{\tilde{A}_k}^+(C_s) + v_{\tilde{A}_k}^+(C_s) \le 3$, $u_{\tilde{A}_k}^-(C_s) \ge 0$, $w_{\tilde{A}_k}^-(C_s) \ge 0$, $v_{\tilde{A}_k}^-(C_s) \ge 0$, $s = 1,2,\ldots,n$ and $k = 1,2,\ldots,m$. The INS value that is the trible of intervals for $C_s$ is denoted by $\alpha_{ks} = ([a_{ks}, b_{ks}], [c_{ks}, d_{ks}], [e_{ks}, f_{ks}])$, where $[a_{ks}, b_{ks}]$ indicates the degree that the alternative $\tilde{A}_k$ satisfies the criterion $C_s$ and $[c_{ks}, d_{ks}]$ indicates the degree that the alternative $\tilde{A}_k$ is indeterminacy on the criterion $C_s$, where as $[e_{ks}, f_{ks}]$ indicates the degree that the alternative $\tilde{A}_k$ does not satisfy the criterion $C_s$ given by the decision-maker. So we can express a decision matrix $= (\tilde{A}_{ks})_{m \times n}$. The aggregating interval neutrosophic number $\alpha_k$ for $\tilde{A}_k$ ($k = 1,2,\ldots,m$) is $\tilde{A}_k = ([a_k, b_k], [c_k, d_k], [e_k, f_k]) = F_{k\omega}(\tilde{A}_{k1}, \tilde{A}_{k2}, \ldots, \tilde{A}_{kn})$ or $\tilde{A}_k = ([a_k, b_k], [c_k, d_k], [e_k, f_k]) = G_{k\omega}(\tilde{A}_{k1}, \tilde{A}_{k2}, \ldots, \tilde{A}_{kn})$, which is obtained by applying *Definition 2.15* or *Definition 2.16* according to each row in the decision matrix.

We can summarize the procedure of proposed method as follows:

**Step (1)** Obtain the weighted arithmetic average values by using Eq. (3) or the weighted geometric average values by Eq. (4).

**Step (2)** Obtain the score (or accuracy) $L(\tilde{A}_k)$ of interval neutrosophic value $\tilde{A}_k$ ($k = 1,2,\ldots,m$) by using Eq. (6).

**Step (3)** Rank the alternative $\tilde{A}_k = (k = 1,2,\ldots,m)$ and choose the best one(s) according to $(\tilde{\alpha}_k)$ ($k = 1,2,\ldots,m$).

## 4.1. Numerical examples

**Example 5.32** Let us consider decision making problem adapted from [32]. There is an investment company, which wants to invest a sum of money in the best option. There is a panel with four possible alternatives to invest the money: (1) $A_1$ is a food company; (2) $A_2$ is a car company; (3) $A_3$ is an arms company; (4) $A_4$ is a computer company. The investment company must make a decision according to three criteria given below: (1) $C_1$ is the growth analysis; (2) $C_2$ is the risk analysis; (3) $C_3$ is the environmental impact analysis. Then, the weight vector of the criteria is given by are $0.35, 0.25$ and $0.40$. Thus, when the four possible alternatives with respect to the above three criteria are evaluated by the expert, we can obtain the following single-valued neutrosophic decision matrix:

|       | $C_1$         | $C_2$         | $C_3$         |
|-------|---------------|---------------|---------------|
| $A_1$ | (0.4,0.2,0.3) | (0.4,0.2,0.3) | (0.2,0.2,0.5) |
| $A_2$ | (0.6,0.1,0.2) | (0.6,0.1,0.2) | (0.5,0.2,0.2) |
| $A_3$ | (0.3,0.2,0.3) | (0.5,0.2,0.3) | (0.5,0.3,0.2) |
| $A_4$ | (0.7,0.0,0.1) | (0.6,0.1,0.2) | (0.4,0.3,0.2) |

Suppose that the weights of $C_1$, $C_2$ and $C_3$ are $0.35, 0.25$ and $0.40$. Then, we use the approach developed to obtain the most desirable alternative(s).

**Step (1)** We can compute the weighted arithmetic average value $\alpha_k$ for $A_k = (k = 1,2,3,4)$ by using Eq. (1) as follows:
$$\alpha_1 = (0.3268, 0.2000, 0.3680),$$
$$\alpha_2 = (0.5626, 0.1319, 0.2000),$$
$$\alpha_3 = (0.4375, 0.2352, 0.2550),$$
$$\alpha_4 = (0.5746, 0.0000, 0.1569).$$

**Step (2)** By using Eq. (5), we obtain $K(\alpha_k)$ $(k = 1,2,3,4)$ as

$$K(\alpha_1) = 0.2794, K(\alpha_2) = 0.5494, K(\alpha_3) = 0.3560,$$
$$K(\alpha_4) = 0.7088.$$

**Step (3)** Rank all alternatives according to the accuracy degrees of $K(\alpha_k)$ $(k = 1,2,3,4)$:

$$A_4 > A_2 > A_3 > A_1.$$

Thus the alternative $A_4$ is the most desirable alternative based weighted arithmetic average operator.

Now, assuming the same weights for $C_1$, $C_2$ and $C_3$, we use the weighted geometric average operator.

**Step (1)** We can obtain the weighted arithmetic average value $\alpha_k$ for $A_k = (k = 1,2,3,4)$ by using Eq. (2) as follows:
$$\alpha_1 = (0.2297, 0.2000, 0.3674),$$
$$\alpha_2 = (0.5102, 0.1860, 0.1614),$$
$$\alpha_3 = (0.3824, 0.2000, 0.2260),$$
$$\alpha_4 = (0.4799, 0.1555, 0.1261).$$

**Step (2)** By applying Eq. (5), we obtain $K(\alpha_k)$ $(k = 1,2,3,4)$ as

$$K(\alpha_1) = 0.2311, K(\alpha_2) = 0.4884, K(\alpha_3) = 0.3782,$$
$$K(\alpha_4) = 0.5412.$$

**Step (3)** Rank all alternatives according to the accuracy degrees of $K(\alpha_k)$ $(k = 1,2,3,4)$:

$$A_4 > A_2 > A_3 > A_1.$$

Thus the alternative $A_4$ is also the most desirable alternative based weighted geometric average operator.

**Example 5.33** Let us consider decision making problem adapted from [30]. Suppose that there is a panel with four possible alternatives to invest the money: (1) $\tilde{A}_1$ is a food company; (2) $\tilde{A}_2$ is a car company; (3) $\tilde{A}_3$ is an arms company; (4) $\tilde{A}_4$ is a computer company. The investment company must make a decision according to three criteria given below: (1) $C_1$ is the growth analysis; (2) $C_2$ is the risk analysis; (3) $C_3$ is the environmental impact analysis. By using the interval-valued intuitionistic fuzzy information, the decision-maker has evaluated the four possible alternatives under the above three criteria and has listed in the following matrix:

|       | $C_1$ | $C_2$ |
|-------|-------|-------|
| $\tilde{A}_1$ | ([0.4,0.5], [0.2,0.3], [0.3,0.4]) | ([0.4,0.6], [0.1,0.3], [0.2,0.4]) |
| $\tilde{A}_2$ | ([0.6,0.7], [0.1,0.2], [0.2,0.3]) | ([0.6,0.7], [0.1,0.2], [0.2,0.3]) |
| $\tilde{A}_3$ | ([0.3,0.6], [0.2,0.3], [0.3,0.4]) | ([0.5,0.6], [0.2,0.3], [0.3,0.4]) |
| $\tilde{A}_4$ | ([0.7,0.8], [0.0,0.1], [0.1,0.2]) | ([0.6,0.7], [0.1,0.2], [0.1,0.3]) |

|       | $C_3$ |
|-------|-------|
| $\tilde{A}_1$ | ([0.7,0.9], [0.2,0.3], [0.4,0.5]) |
| $\tilde{A}_2$ | ([0.3,0.6], [0.3,0.5], [0.8,0.9]) |
| $\tilde{A}_3$ | ([0.4,0.5], [0.2,0.4], [0.7,0.9]) |
| $\tilde{A}_4$ | ([0.6,0.7], [0.3,0.4], [0.8,0.9]) |

Suppose that the weights of $C_1$, $C_2$ and $C_3$ are $0.35, 0.25$ and $0.40$. Then, we use the approach developed to obtain the most desirable alternative(s).

**Step (1)** We can compute the weighted arithmetic average value $\tilde{\alpha}_k$ for $\tilde{A}_k = (k = 1,2,3,4)$ by using Eq. (4) as follows:

$\tilde{\alpha}_1 = ([0.5452, 0.7516], [0.1681, 0.3000], [0.3041, 0.4373])$,
$\tilde{\alpha}_2 = ([0.4996, 0.6634], [0.1551, 0.2885], [0.3482, 0.4655])$,
$\tilde{\alpha}_3 = ([0.3946, 0.5626], [0.2000, 0.3365], [0.4210, 0.5532])$,
$\tilde{\alpha}_4 = ([0.6383, 0.7396], [0.0000, 0.2070], [0.2297, 0.4039])$.

**Step (2)** By using Eq. (6), we obtain $L(\tilde{\alpha}_k)$ $(k = 1,2,3,4)$ as

$$L(\tilde{\alpha}_1) = 0.4048, L(\tilde{\alpha}_2) = 0.3655, L(\tilde{\alpha}_3) = 0.2275,$$
$$L(\tilde{\alpha}_4) = 0.5825.$$

**Step (3)** Rank all alternatives according to the accuracy degrees of $L(\tilde{\alpha}_k)$ $(k = 1,2,3,4)$:

$$\tilde{A}_4 > \tilde{A}_1 > \tilde{A}_2 > \tilde{A}_3.$$

Thus the alternative $\tilde{A}_4$ is the most desirable alternative based weighted arithmetic average operator.

Now, assuming the same weights for $C_1$, $C_2$ and $C_3$, we use the weighted geometric average operator.

**Step (1)** We can obtain the weighted arithmetic average value $\tilde{\alpha}_k$ for $\tilde{A}_k =$ $(k = 1,2,3,4)$ by using Eq. (4) as follows:

$\tilde{\alpha}_1 = ([0.5003, 0.6620], [0.1760, 0.3000], [0.3195, 0.4422])$,
$\tilde{\alpha}_2 = ([0.4547, 0.6581], [0.1860, 0.3371], [0.5405, 0.6758])$,
$\tilde{\alpha}_3 = ([0.3824, 0.5578], [0.2000, 0.3418], [0.5012, 0.7069])$,
$\tilde{\alpha}_4 = ([0.6332, 0.7334], [0.1555, 0.2569], [0.5068, 0.6632])$.

**Step (2)** By applying Eq. (6), we obtain $L(\tilde{\alpha}_k)$ $(k = 1,2,3,4)$ as

$$L(\tilde{\alpha}_1) = 0.3621, L(\tilde{\alpha}_2) = 0.2118, L(\tilde{\alpha}_3) = 0.1621,$$
$$L(\tilde{\alpha}_4) = 0.3429.$$

**Step (3)** Rank all alternatives according to the accuracy degrees of $L(\tilde{\alpha}_k)$ $(k = 1,2,3,4)$:

$$\tilde{A}_1 > \tilde{A}_4 > \tilde{A}_2 > \tilde{A}_3.$$

Thus the alternative $\tilde{A}_1$ is also the most desirable alternative based weighted geometric average operator.

Note that we obtain the different rankings for single valued neutrosophic information and interval neutrosophic information.

From the examples, we can see that the proposed neutrosophic decision-making method is more suitable for real scientific and engineering applications because it can handle not only incomplete information but also the indeterminate information and inconsistent information existing in real situations. The technique proposed in this paper extends the existing decision making methods and provides a new way for decision makers.

## 6. Comparison Analysis and Discussion

In this section, we will a comparison analysis to validate the feasibility of the proposed decision making method based on accuracy-score functions. To demonstrate the relationships, we utilize the same examples adapted from [32] and [30].

The score and accuracy functions has extremely important for process of multi criteria decision making. But, until now there have been no many studies on multi-criteria decision making method based on accuracy-score functions, which are criterion values for alternatives are single valued neutrosophic sets or interval neutrosophic sets. Ye [30] defined the similarity measures between INSs based on the relationship between similarity measures and distances and proposed the similarity measures between each alternative and the ideal alternative to establish a multi criteria decision making method for INSs. After, Zhang et al. [6] presented a method based on the aggregation operators for multi criteria decision making under interval neutrosophic environment. By obtaining the different results than given in [30], they showed that the method proposed is more precise and reliable than the result produced in [30]. Although the same ranking results with [6] are obtained in here, the decision making method proposed in this paper has less calculation and it is more flexible and more sustainable for the multi criteria decision making with SVN or IVN information.

## 7. Conclusions

At present, many score-accuracy function technical are applied to the problems based on intuitionistic fuzzy information or interval valued intuitionistic fuzzy information, but they could not be used to handle the problems based on neutrosophic information. So, two measurement functions such that score and accuracy functions for single valued neutrosophic numbers and interval neutrosophic numbers is proposed in this paper, and a multi-criteria decision making method based on this functions is established for neutrosophic information. In decision making process, the neutrosophic weighted aggregation operators (arithmetic and geometric average operators) are adopted to aggregate the neutrosophic information related to each alternative. Finally, some numerical examples are presented to illustrate the application of the proposed approaches.